\documentclass[11pt]{article}
\usepackage{eamt24}
\usepackage{times}
\usepackage{url}
\usepackage{latexsym}
\usepackage[small,bf]{caption} 
\usepackage{hyperref}
\usepackage{times}
\usepackage{comment}
\usepackage{latexsym}
\usepackage{graphicx}
\usepackage[T1]{fontenc}

\usepackage[utf8]{inputenc}

\usepackage{microtype}

\usepackage{inconsolata}

\usepackage{booktabs}
\usepackage{caption}
\usepackage{subcaption}

\title{{MTUncertainty}: Assessing the Need for Post-editing \\of Machine Translation Outputs by Fine-tuning OpenAI LLMs}

\author{  Serge Gladkoff$ ^2$,
          Lifeng Han$ ^{*1}$, Gleb Erofeev$^{ 2}$,    Irina Sorokina$^{ 2}$,  
\and \textbf{Goran Nenadic}$^{1}$ \\
         $^1$ The University of Manchester, UK \\ 
         $^2$ Logrus Global,  Translation \& Localization 
         \\ {\tt lifeng.han, g.nenadic @ manchester.ac.uk} 
         \\
         {\tt 
         gleberof, irina.sorokina, serge.gladkoff @ logrusglobal.com} 
        \\ $^* corresponding \, author$}

\date{}

\begin{document}
\maketitle

\begin{abstract}
Translation Quality Evaluation (TQE) is an essential step of the modern translation production process. TQE is critical in assessing both machine translation (MT) and human translation (HT) quality without reference translations. The ability to evaluate or even simply estimate the quality of translation automatically may open significant efficiency gains through process optimisation.
This work examines whether the state-of-the-art large language models (LLMs) can be used for this uncertainty estimation of MT output quality. We take OpenAI models as an example technology and approach TQE as a binary classification task.
On \textbf{eight language pairs} including English to Italian, German, French, Japanese, Dutch, Portuguese, Turkish, and Chinese, our experimental results show that fine-tuned \textbf{\textit{GPT3.5}} can demonstrate good performance on translation quality prediction tasks, i.e. \textit{whether the translation needs to be edited}.
Another finding is that simply increasing the sizes of LLMs does not lead to apparent better performances on this task by comparing the performance of three different versions of OpenAI models: \textbf{\textit{Curie}}, \textbf{\textit{Davinci}}, and \textbf{\textit{GPT3.5}} with 13B, 175B, and 175B parameters, respectively. 

\end{abstract}


\section{Introduction}
\label{sec_intro}
Most modern translation projects include post-editing (PE) of machine-translation (MT) output \cite{han_gladkoff_metaeval_tutorial2022,gladkoff-etal-2022-measuring}. Instead of translating from scratch, the MT+PE process increases productivity and allows to speed up global content delivery \cite{gladkoff-han-2022-hope,han-etal-2013-quality}.
However, in regulated industries and many other scenarios raw MT output is not suitable for final publication due to the inevitable errors caused by the inherently stochastic nature of neural MT (NMT) \cite{han2022investigation,google2021human_evaluation_TQA,hong2024cantonmt}. 
Hallucinations, incorrect terminology, factual and accuracy errors, small and large, as well as many other types of mistakes are inevitable to varying degrees of extent, and therefore for premium quality publication human revision is required.
MT output serves as input for a professional human translator, who reviews and revises the MT proposals to eliminate factual errors and ensure that the quality of translated material conforms to the customer specifications.
At the same time even with those languages that are not handled well by MT, there is a significant portion of segments that are not changed after human review. 
This portion varies from 10\% to 70\% in some cases \footnote{\url{logrusglobal.com} statistics}, and the question arises, ``Is it possible to use machine learning (ML) methods to mark these segments and save time for human reviser and make them focus on those segments that need attention instead''? In other words, \textit{Is it possible to capture editing distance patterns from data of prior editing of this material, which already has been made}?
This could further speed up the translation process and decrease the costs while preserving the premium quality of the translated product.

This problem is also closely related to the traditional MT quality estimation (QE) shared task that has been held with the Workshop of MT (WMT) series since 2012 \cite{ws-WMT2012-statistical,wmt-2022-machine,zerva-etal-wmtQE2022-findings,han-etal-2013-quality,han2022overview_mte}, where both token-level and segment-level QE were carried out.

From practical application and industrial usage, we formulate the problem into a single classification task, i.e. we are trying to solve a classification task to answer if the translated segment (sentence) needs to be edited, or not.

With the development of current large language models (LLMs), we choose OpenAI models as state-of-the-art LLMs to examine their capabilities for this task.
In this work, our first experimental investigation is on ``\textbf{Predictive Data Analytics with AI: assessing the need for post-editing of MT output by fine-tuning OpenAI LLMs}''.
We also follow up with an experiment that explores ``\textbf{if the size of sample or LLM matters in such a task}'' by experimenting with three OpenAI models: \textbf{\textit{curie}}, \textbf{\textit{davinci}}, and \textbf{\textit{gpt3.5}}, with parameter sizes varying from 13B to 175B.

The rest of this paper is designed as below. Section \ref{sec_related} introduces related work to ours including MT-QE-related shared task and challenge events, Section \ref{sec_method_pilot} presents our methodology design and pilot study using two language pairs, Section \ref{sec_experiment_extend} extends the experimental investigation with six more language pairs,  
section \ref{sec_japanesenews} discusses experiment on English-Japanese news content with the increasing sizes of training and testing corpus and explores two more OpenAI LLMs with varying model sizes, and Section \ref{sec_conclude} concludes this paper with future work and research perspectives.

\section{Related Work}
\label{sec_related}
The Quality Evaluation (QE) of MT output has always been a critical topic for MT development due to its critical role in assessing quality in the process of training. In many cases, evaluation has to be done without seeing the reference translations. In many practical situations, reference translations are not available or even impossible to acquire, i.e. it is not practical to ``manufacture'' them for evaluation.
The earliest QE shared task with the annual WMT conference started in 2012 when word level QE was introduced by \cite{ws-WMT2012-statistical} to estimate if the translated tokens need to be edited or not, such as deletion, substitution, or keeping it as it is. 
In the later development of QE, a sentence-level task was introduced to predict the overall segment translation scores, which are to be correlated with human judgement scores, such as using Direct Assessment \cite{DBLP:conf/naacl/GrahamBM15}. 
In WMT-2022, a new task on binary sentence-level classification was also introduced to predict if a translated output has critical errors to be fixed on English-German and Portuguses-English language pairs \cite{zerva-etal-wmtQE2022-findings}.

The recent methods used for such QE tasks included prompt-based learning using  XLM-R by KU X Upstage (Korea University, Korea \& Upstage) from \newcite{eo-etal-QE2022-ku}, 
Direct Assessment and MQM features integration into fine-tuning on XLM-R and \textsc{Info}XLM \cite{chi-etal-2021-infoxlm} by the Alibaba team \cite{bao-etal-2022-alibaba}, and
incorporating a word-level sentence tagger and explanation extractor on top of the COMET framework by \newcite{rei-etal-2022-cometkiwi},
in addition to historical statistical methods such as support vector machine (SVM), Naive Bayes classifier (NB), and Conditional Random Fields (CRFs) by \newcite{han-etal-2013-quality}.

However, \textit{to the best of our knowledge}, this work is the first 
to investigate the OpenAI LLMs with varying sizes on such MT error prediction tasks with positive outcomes.


\section{Methodology and Experiments}
\label{sec_method_pilot}

\begin{figure}[!t]
\begin{center}
\centering
\includegraphics*[width=0.48\textwidth]{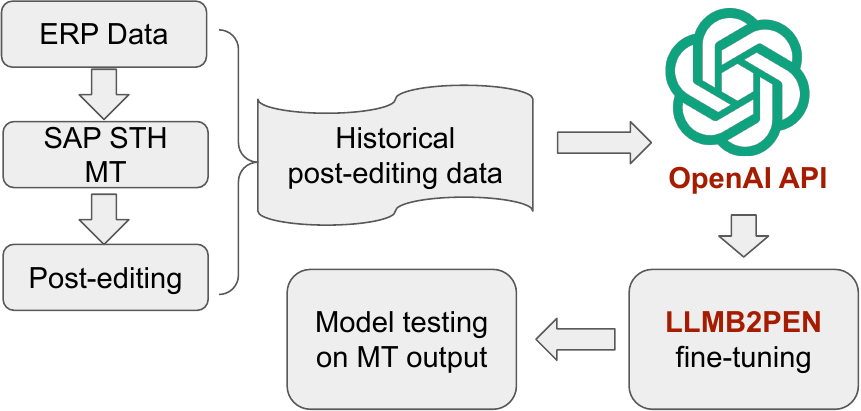}
\caption{LLMB2PEN Methodology Design on Fine-tuning LLMs for Binary Prediction of Post-editing Need on Translations. 
} 
\label{fig:chatGPT-finetune-diagram}
\end{center}
\end{figure}

\begin{figure*}[!t]
\begin{center}
\centering
\includegraphics*[width=0.99\textwidth]{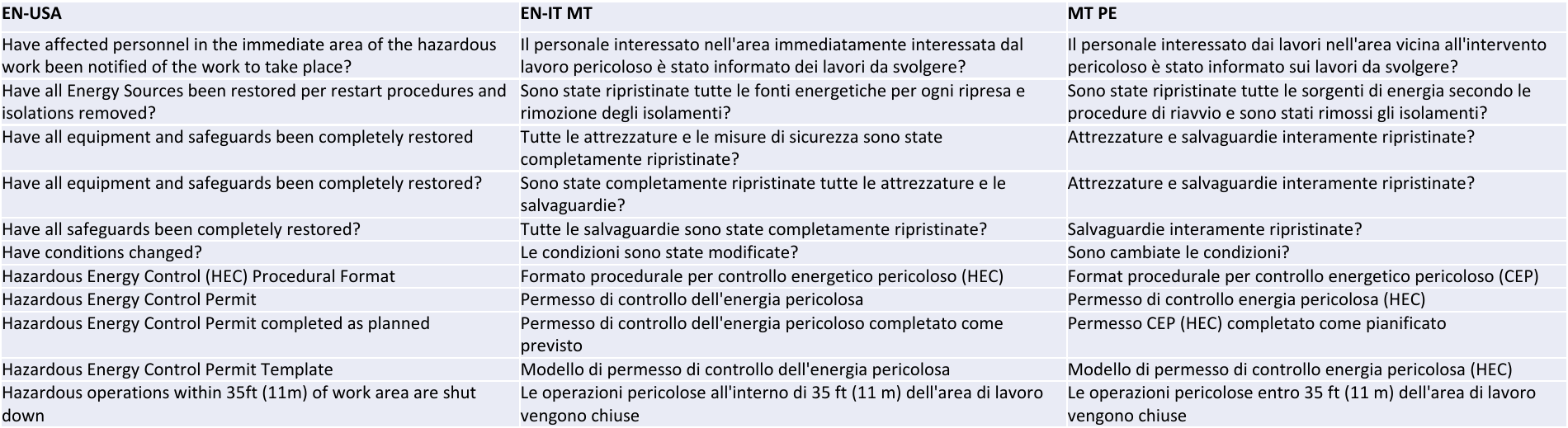}
\caption{EN-IT Examples on MT and Post-Editing}
\label{fig:en-it-MTPE-example}
\end{center}
\end{figure*}

\begin{figure*}[!t]
\begin{center}
\centering
\includegraphics*[width=0.99\textwidth]{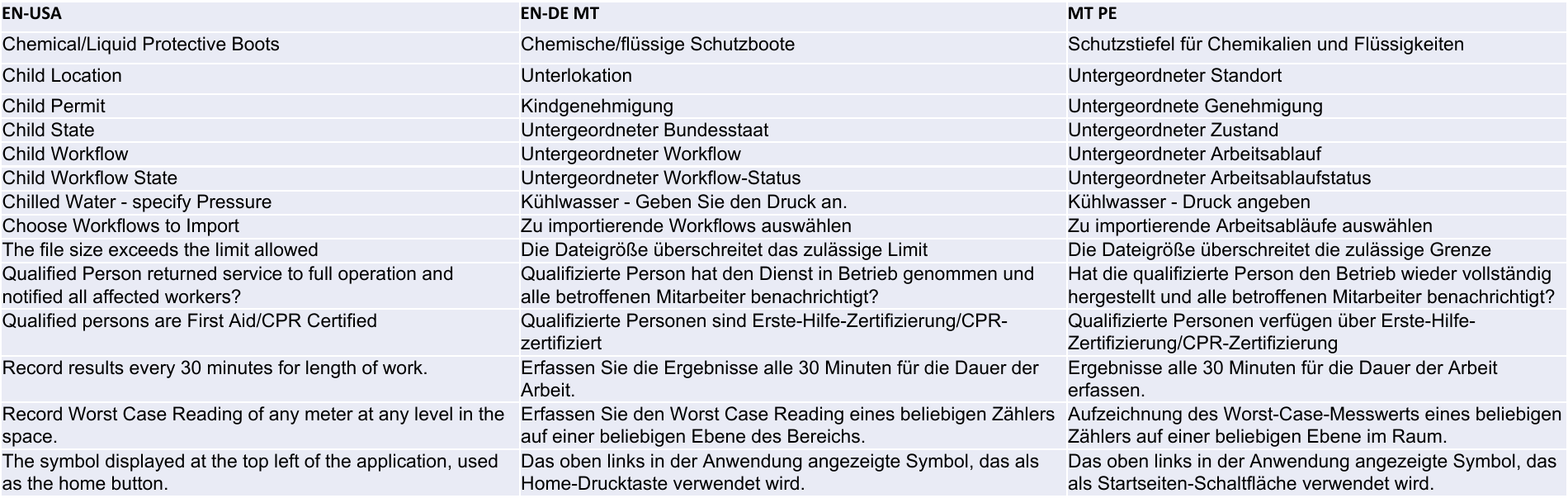}
\caption{EN-DE Examples on MT and Post-Editing}
\label{fig:en-de-MTPE-example}
\end{center}
\end{figure*}

As shown in the system diagram in Figure \ref{fig:chatGPT-finetune-diagram}, we first collect the historical post-editing data from our past projects on eight languages of Enterprise Resource Planning (ERP) content translation on English$\rightarrow$German, French, Italian, Japanese, Dutch, Portuguese, Turkish, and Chinese (DE, FR, IT, JA, NL, PT, TR, ZH). %
This project was completed by using an MT engine to automatically translate the source into the eight languages, followed by post-editing by professional linguists. Two examples of MT and PE in English-Italian and English-German languages as Pilot Experiments are shown in Figure \ref{fig:en-it-MTPE-example} and \ref{fig:en-de-MTPE-example}.
Regarding MT system selection, since the content was from the ERP domain, we used the SAP STH as our MT engine. \footnote{\url{https://www.sap.com/} SAP is an enterprise resource planning, automation and business software company.}


With this data from a real-world translation project, we used API to fine-tune the OpenAI \textbf{\textit{curie}} model for our classification task. The input is the triple set (English source, MT outputs, post-edited "gold standard") we prepared in Phase 1.
The goal of this step is to optimise the weights of the model parameters for our classification task.
The custom fine-tuned model 
produced as a result of LLMB2PEN (LLM for Binary Prediction of Post-editing Need) method is created in our private space on the OpenAI account.

We \textit{did not} apply ``prompt engineering'' for this task by doing zero-shot, one-shot, or few-shot training; we \textit{did a full-scale fine-tuning} of OpenAI LLMs via API. It is important to note that we did not simply train the LLM for edit distance either; instead, the model was trained to learn whether the strings were edited or not taking into account the full content of the string and the entire context of the training data.
One of the reasons that we did not use prompting is that 
``Prompt Engineering'' of ChatGPT-3 is limited by 3,000 tokens, and with ChatGPT-4 the context has been increased to 25,000 tokens, but still very significant limitation remains. OpenAI documentation states that 100 tokens = 75 words, meaning that the average sentence is 20 tokens, therefore 3000 tokens is only 150 sentences, or 75 translation units of bilingual text, or 50 segment triples of source, target and reference. The context of 25,000 sentences is only about 150 segment triples.

Also, fine-tuning is a deeper process of adjusting the model's weights, and not just an in-context learning. That's why we chose fine-tuning method, which is not constrained by such limitations.

For our classification experiment we took about 4000 lines of bilingual data in triples of source, target, and reference, and split it into train (large) and test (smaller) sets with a ratio of 9:1. 

There were no specific selection criteria for the data because we took the entire project dataset after project completion. (Please, note that since we used the entire data from the actual project, and split the data set as 9:1, the sizes of test sets are not round and slightly different for different languages.)

We also combined source sentences in groups of length, so that the test data set has the same distribution of sentences by their length as the training dataset.

Since the average sentence size is about 17 words, the training dataset contained about 35000 words of source data, 35000 words of MT output, and 35000 words of post-edited human reference.

It is also important to note what the model learns in this case - in such an experiment it learns not to translate, but to spot MT translation errors that were made by the specific MT engine in a specific language pair on particular content.

\subsection{Outputs on EN-DE/IT}

\begin{figure}[!t]
\begin{center}
\centering
\includegraphics*[width=0.48\textwidth]{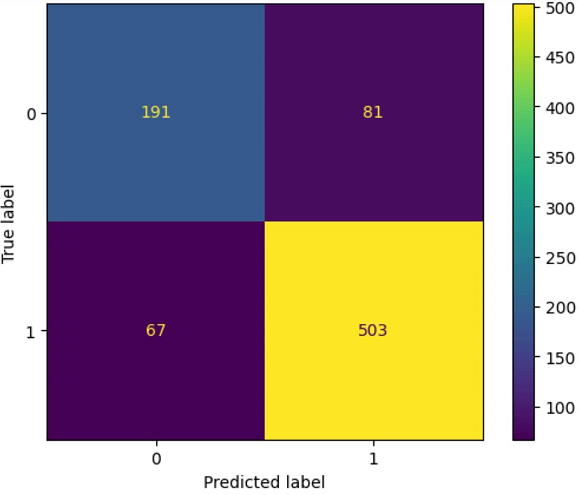}
\caption{EN-IT Confusion Matrix of LLMB2PEN, \textbf{\textit{curie}} model: Clockwise from top-left corner (TN, FP, TP, FN)}
\label{fig:en-it-confusion}
\end{center}
\end{figure}

\begin{figure}[!t]
\begin{center}
\centering
\includegraphics*[width=0.48\textwidth]{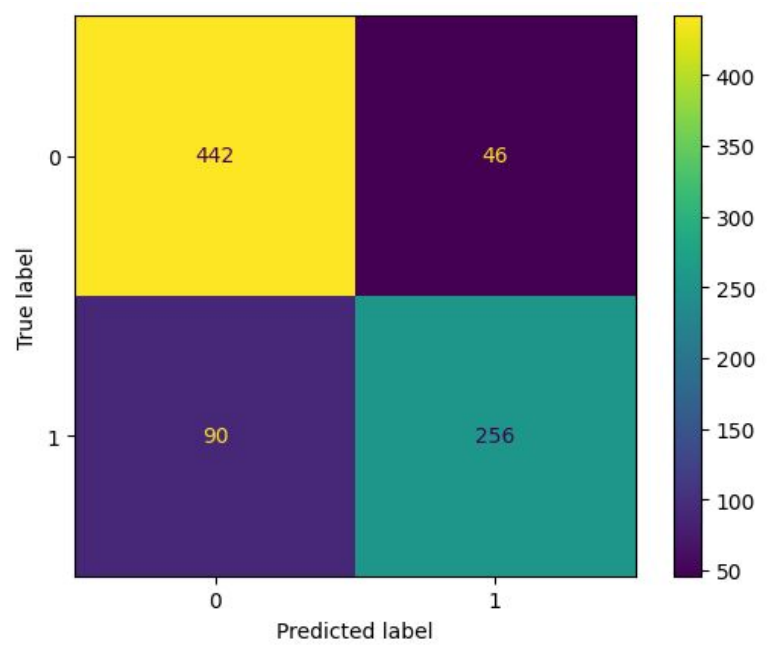}
\caption{EN-DE Confusion Matrix of LLMB2PEN, \textbf{\textit{curie}} model: Clockwise from top-left corner (TN, FP, TP, FN)}
\label{fig:en-de-confusion}
\end{center}
\end{figure}

As a first step, we trained the \textbf{\textit{curie}} LLM model using our data for two language pairs: English-Italian and English-German.
To illustrate the results of prediction with our LLMB2PEN method, we draw the confusion matrix for both language pairs in Figures \ref{fig:en-it-confusion} and \ref{fig:en-de-confusion}. 

In the Confusion Matrix, from the top left corner in a clockwise direction, the 1st quadrant means True Negative (\textbf{TN}): segment is predicted as not requiring editing and it does not indeed require post-editing. 
The 2nd quadrant is False Positive (FP): segments which are predicted as requiring editing, but in reality, they do not, that is \textbf{FP} means that the segment is correct but wrongly flagged for post-editing.  
The 3rd quadrant is True Positive (\textbf{TP}) - reflecting the situation when a segment is correctly flagged as requiring post-editing.
The fourth quadrant is False Negative (\textbf{FN}): segment is predicted as correct, while in reality, it does require post-editing.
So the first and third are successful classifications, and the other two are incorrect classifications.

It is worth mentioning that if the segment is incorrectly predicted as requiring post-editing, this only leads to a small increase in post-editing cost, while False Negative predictions represent the consumer's risk of seeing substandard segments as not corrected in the final product. So in the context of our task, we are much more concerned with the share of False Negatives in the test classification dataset.

In the Italian situation shown in Figure \ref{fig:en-it-confusion}, you can see that the model predicts correctly that many more translated sentences need to be edited (TP=503) than sentences that do not need to be edited (TN=191). In incorrectly predicted categories, 67 sentences need to be edited but predicted as good, and 81 translated sentences do not need to be edited, but the prediction says they have to be reviewed.

In the English-German set from Figure \ref{fig:en-de-confusion}, the situation is the opposite:  there are more translated sentences that do not need to be edited (442) than prescribed for review (256) in the correct predictions. In the wrong prediction categories, such numbers are 90 and 46 respectively.

The prediction \textbf{accuracy} of the LLMB2PEN model on our designed task is \textbf{(TP+TN)/Total} = (503+191)/842 = 82.42\% for English-Italian MT, and 
(442+256)/834 = 83.69\% for English-German MT.
Overall, our LLMB2PEN method shows that the English-German output is clearly better than the English-Italian. 

However, if we only count the Type II errors (incorrect prediction that the segments should NOT be edited), then the corresponding error rates will be 67/842 = 8\% for Italian and 90/834 = 10\% for German.

\subsection{Discussion}

The first and foremost finding is that the fine-tuned model learned enough information to make a very significant prediction of whether the segment has to be edited or not.
It should be noted that such successful classification holds the promise of a viable method to significantly reduce the volume of post-editing efforts and therefore time and costs.
There is, however, a problem: while it is OK to present the editor with segments that are predicted as required for editing, but in reality do not require editing (the fourth quadrant, FP), real consumer risk comes from the segments that have been predicted as not requiring editing and made their way to the final predict, but in reality, they contain errors (the fourth quadrant, FN).

Such segments represent a significant portion of segments predicted as not requiring post-editing: \textbf{FN/(TN+FN)} = 67/(191+67) = 67/258 = 26\% of “leave as is” (let’s call them “LAI”) segments for Italian, and 90/(442+90) = 90/532 = 16.9\% for German.

It is possible that for specific language pairs and MT engines the portion of the LAI segments will decrease with the size increase of the training data and further fine-tuning, but it is unlikely to become zero, since with neural models the error rate is never zero.

Two strategies can be considered for implementing such prediction in production:

\begin{enumerate}
    \item The LAI segments are excluded from the human loop and go into publication unvetted, but not straight away
    as they advance through the workflow along with all the other segments. In this scenario, the potential error rate ceiling for final content will be \textbf{FN/Total} = FN/(TP+FN+TN+FP) 
    = 8\% for Italian, i.e. 67/(81 + 67 + 191 + 503) = 81/842 and 10.8\%= 90 / (90 + 46+ 442 + 256) = 90/834 for German.
    
    It is not impossible to predict what would be the actual error rate in those 8\% and 10.8\% segments that will not be reviewed or the severity of errors in them. It is, obviously, the decision of the customer to decide whether this is an acceptable level of consumer risk for their situation (domain, type of content, audience, etc.). Additional risk assessment may be required to be carried out.
    
    The savings on post-editing volume in this scenario would be (TN+FN)/Total = (191+67)/842 = 30.1\% for Italian and (442+90)/834 = 63.8\% for German.
    
    \item All LAI segments are marked as ``100\% MT matches'' in a CAT tool. With this approach, translators are requested to review them, but at a lower per-word rate, using the traditional approach which is well familiar to translation providers. In this scenario the reduction of the total time, effort, and cost can be estimated as follows: without this approach, translators working on the Edit Distance Calculation (EDC) model will get lower payment (which can vary from 10\% to 40\% with different payment models) for not changed segments. In this scenario, translators may be asked to review such LAI segments but paid only a small part of the full rate for the review of such segments.
\end{enumerate}

Simple proportion allows us to calculate the savings in the second scenario: if we take the full payment for all the segments for 100\% of post-editing costs, and assume that 10\% pay reflects adequate pay for the review of LAI segments that are marked as such, the volume of post-editing decreases 27.6\% for Italian and 57.4\% for German with zero error rate of the final product (no producer’s or consumer’s risk).

This estimate of a potential economy with a guarantee of zero error rate begs for further research and implementation of this method.

\section{Extended Experiments On Six More Language Pairs}
\label{sec_experiment_extend}

\begin{figure}[!h]
\begin{center}
\centering
\includegraphics*[width=0.48\textwidth]{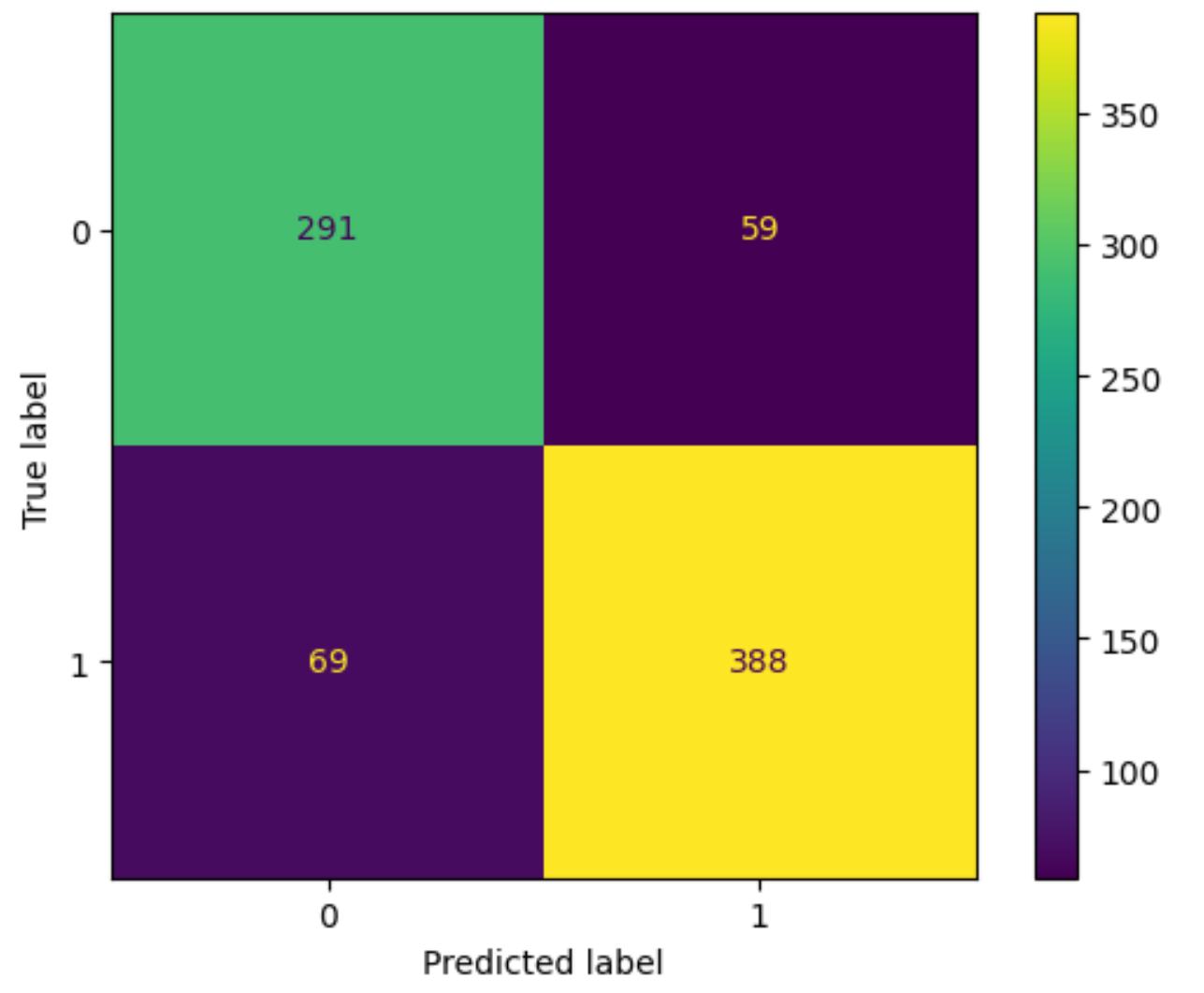}
\caption{EN-FR Confusion Matrix of LLMB2PEN, \textbf{\textit{curie}} model: Clockwise from top-left corner (TN, FP, TP, FN)}
\label{fig:en-fr-confusion}
\end{center}
\end{figure}
\begin{figure}[!h]
\begin{center}
\centering
\includegraphics*[width=0.48\textwidth]{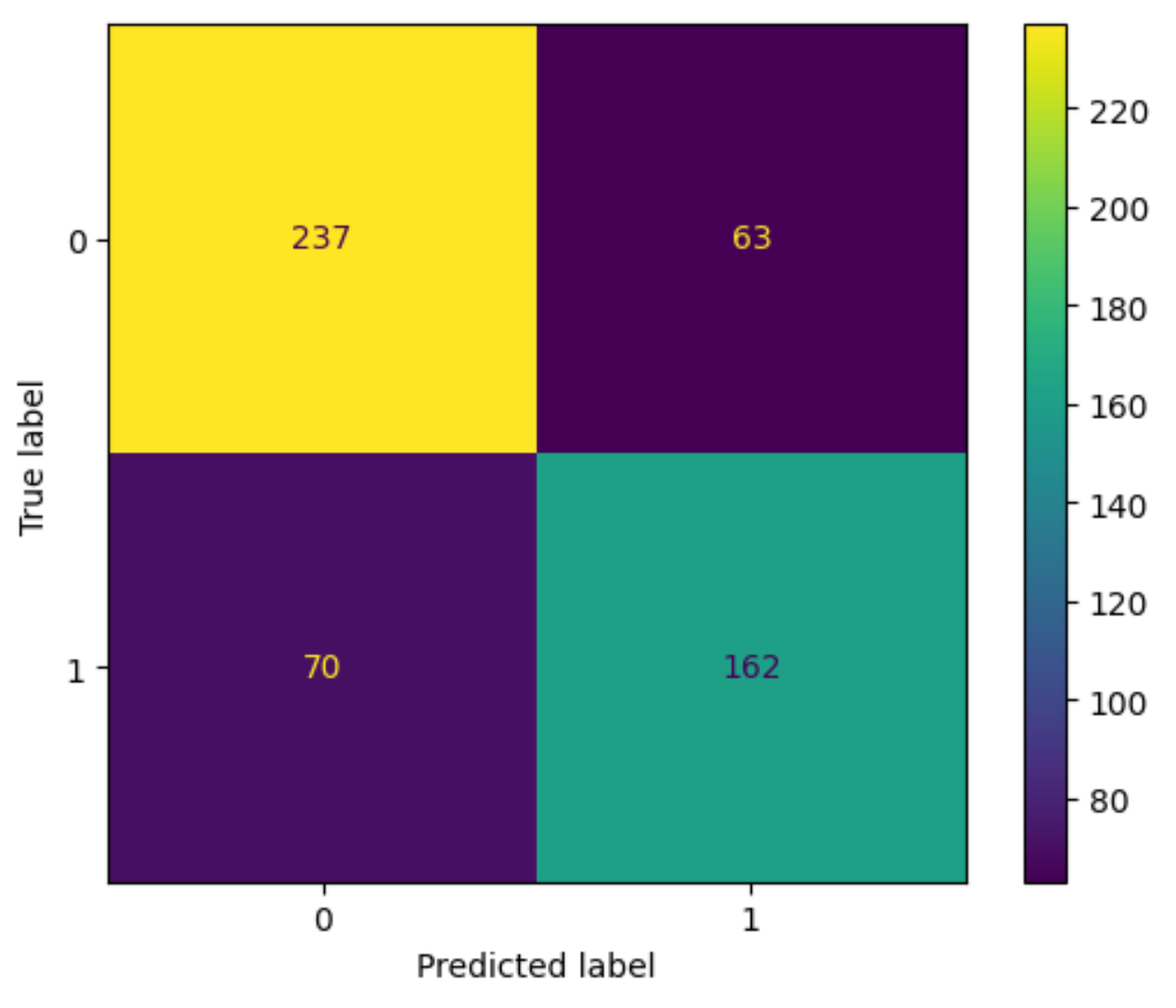}
\caption{EN-JA Confusion Matrix of LLMB2PEN, \textbf{\textit{curie}} model: Clockwise from left-up corner (TN, FP, TP, FN)}
\label{fig:en-ja-confusion}
\end{center}
\end{figure}
\begin{figure}[!h]
\begin{center}
\centering
\includegraphics*[width=0.48\textwidth]{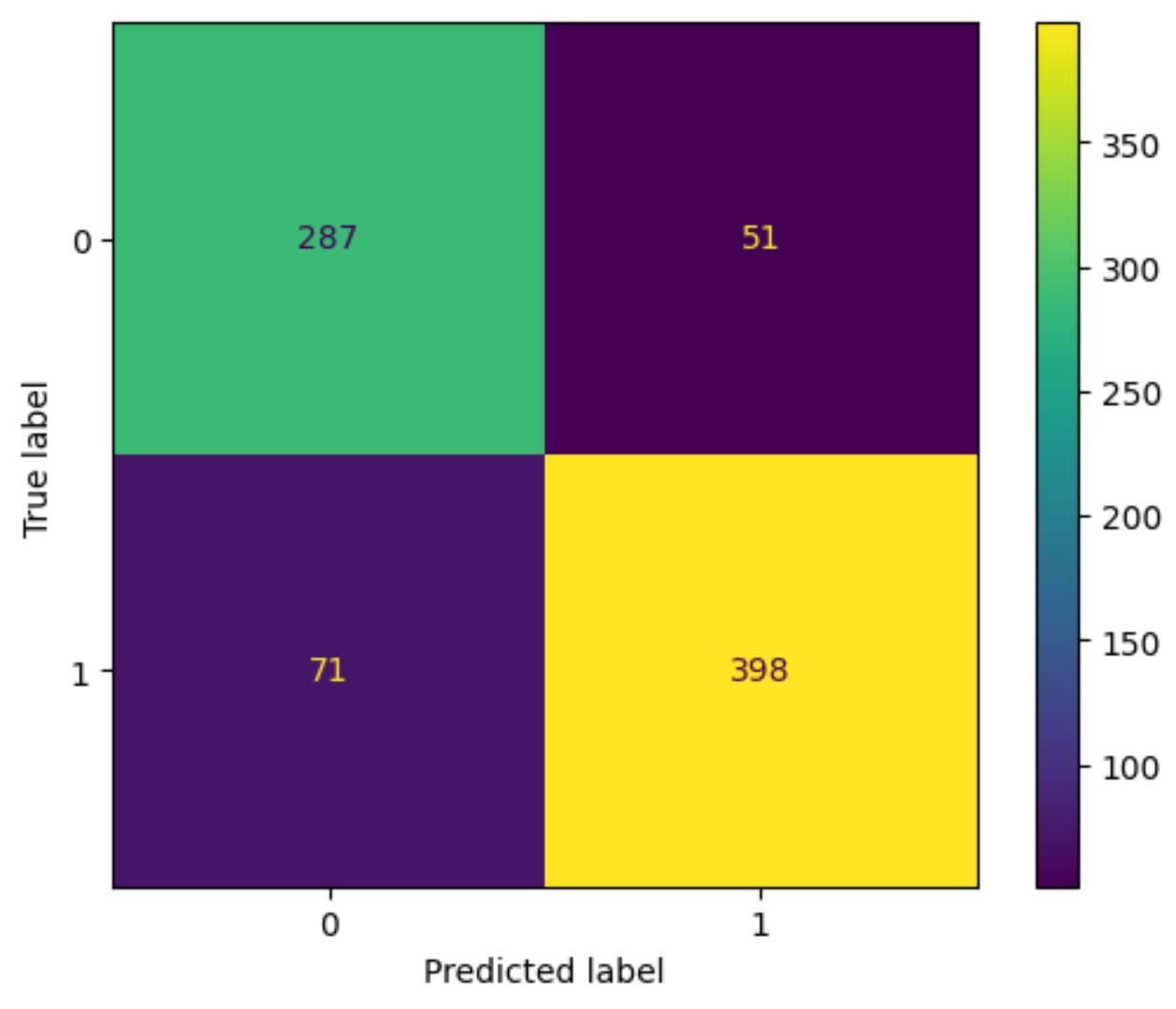}
\caption{EN-NL Confusion Matrix of LLMB2PEN, \textbf{\textit{curie}} model: Clockwise from top-left corner (TN, FP, TP, FN)}
\label{fig:en-NL-confusion}
\end{center}
\end{figure}
\begin{figure}[!h]
\begin{center}
\centering
\includegraphics*[width=0.48\textwidth]{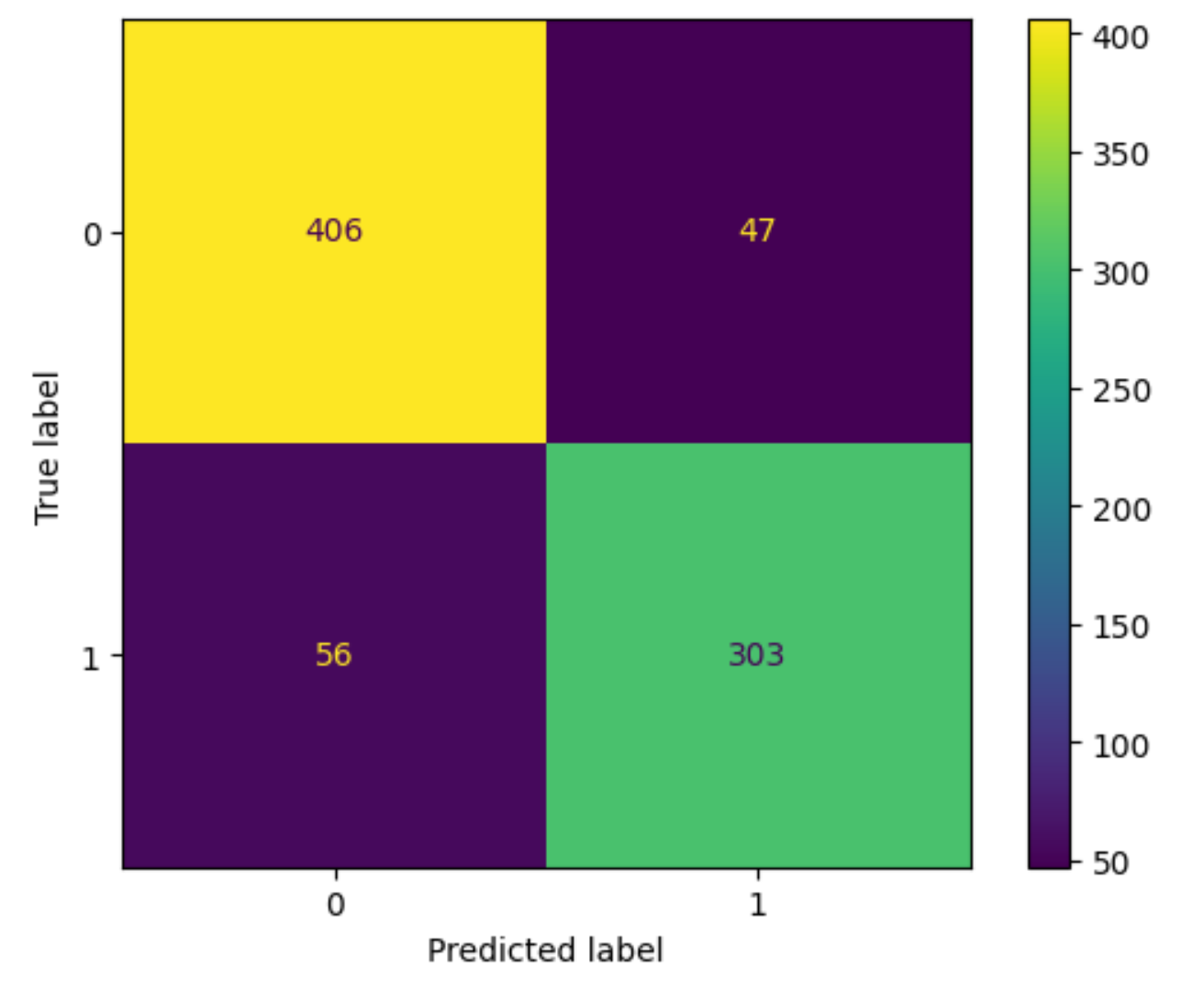}
\caption{EN-PT Confusion Matrix of LLMB2PEN, \textbf{\textit{curie}} model: Clockwise from left-up corner (TN, FP, TP, FN)}
\label{fig:en-PT-confusion}
\end{center}
\end{figure}
\begin{figure}[!h]
\begin{center}
\centering
\includegraphics*[width=0.48\textwidth]{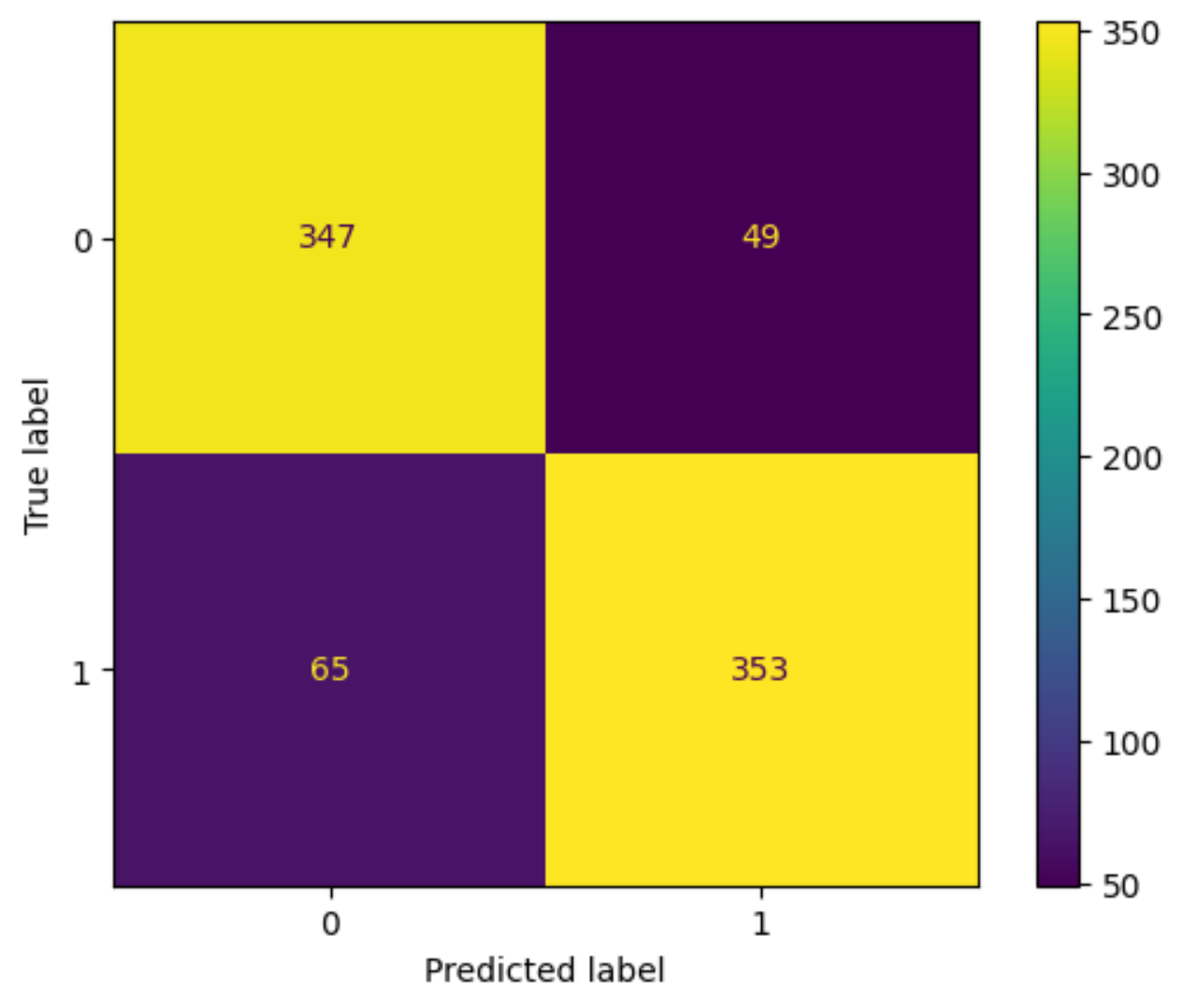}
\caption{EN-TR Confusion Matrix of LLMB2PEN, \textbf{\textit{curie}} model: Clockwise from top-left corner (TN, FP, TP, FN)}
\label{fig:en-TR-confusion}
\end{center}
\end{figure}
\begin{figure}[!h]
\begin{center}
\centering
\includegraphics*[width=0.48\textwidth]{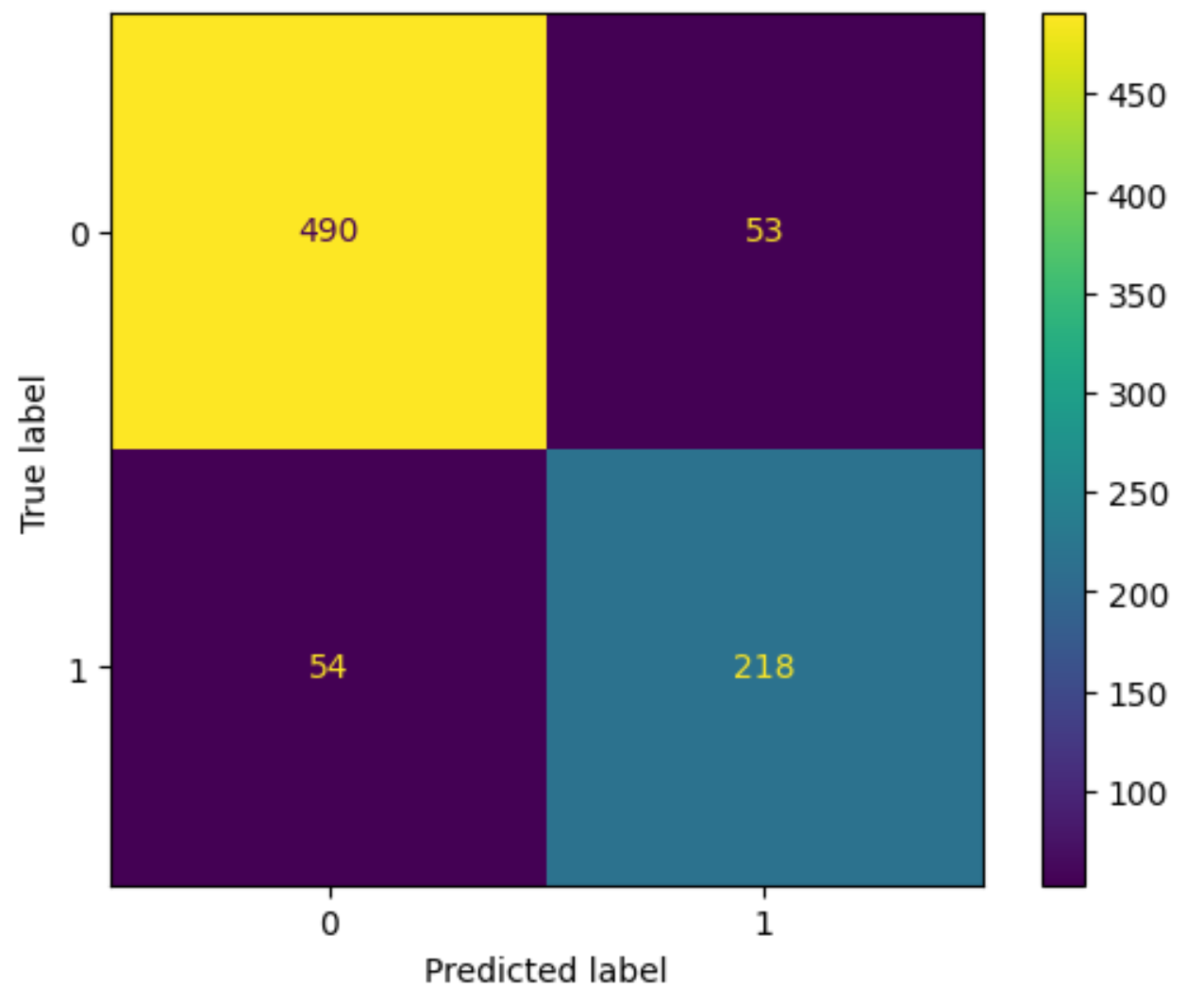}
\caption{EN-ZH Confusion Matrix of LLMB2PEN, \textbf{\textit{curie}} model: Clockwise from top-left corner (TN, FP, TP, FN)}
\label{fig:en-ZH-confusion}
\end{center}
\end{figure}

We hereby also present extended experimental results using six more language pairs obtained with LMB2PEN method for translation editing distance prediction. These language pairs include English-to-French, Japanese, Dutch, Portuguese, Turkish, and Chinese (EN$\rightarrow$FR/JA/NL/PT/TR/ZH), whose results are listed in Figure \ref{fig:en-fr-confusion},  \ref{fig:en-ja-confusion}, \ref{fig:en-NL-confusion}, \ref{fig:en-PT-confusion}, \ref{fig:en-TR-confusion}, and \ref{fig:en-ZH-confusion}
  respectively. 
  
  From the results presented in the figures, in general, the ratio of correct prediction (TP+TN) is much higher than the one from mis-prediction (FN+FP) across all these language pairs, as for English-Italian and English-German in the pilot studies.
  On one hand, the following language pairs have more True Positive than True Negative predicted segments \textbf{than for English-German/Italian}: English-Japanese, English-Portuguese, and English-Chinese.  
  On the other hand, the rest of the language pairs have more TN than TP: English-French, and English-Dutch, except for English-Turkish which has a comparable number of segments between TP (347) and TN (353) labels. 
  This finding also indicates that such language pairs with a high number of TN labels are still much more challenging for MT system development to produce more correct outputs, i.e., English to French, Dutch, and Turkish.
  Earlier research findings from \newcite{gladkoff-etal-2022-measuring} on TQE conclude that 200+ segments can be enough amount of data to reflect the MT system quality.

\section{Different LLMs on EN-JA News Domain}
\label{sec_japanesenews}

In the subsequent experiment on data, we used different news items translation corpus from different projects, translated from English to Japanese.
\subsection{Using OpenAI GPT3.5turbo}
In this experiment, we have repeated experiments of fine-tuning the OpenAI \textbf{\textit{gpt3.5turbo}} model on datasets of different sizes: 2000 pairs, 4000 pairs, and 6000 pairs.

\begin{figure}[!h]
\begin{center}
\centering
\includegraphics*[width=0.45\textwidth]{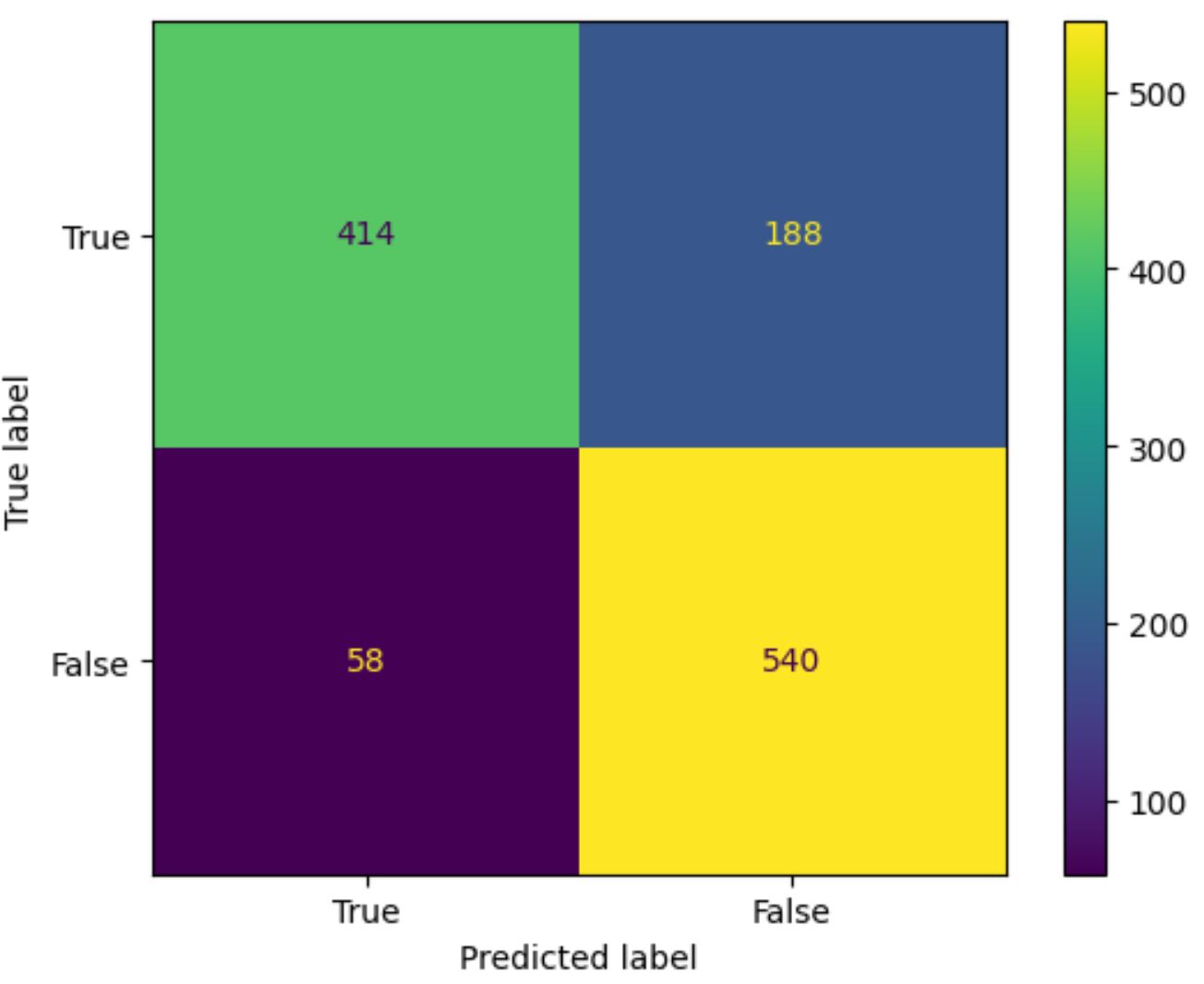}
\caption{EN-JA news items Confusion Matrix of LLMB2PEN, \textbf{\textit{gpt3.5turbo}} model: Clockwise from top-left corner (TP, FN, TN, FP)}
\label{fig:en-JA-news}
\end{center}
\end{figure}

Figure \ref{fig:en-JA-news} shows the confusion matrix for the training set of 6000 bilingual EN-JA translation pairs in the news domain.

\begin{figure}[!h]
\begin{center}
\centering
\includegraphics*[width=0.48\textwidth]{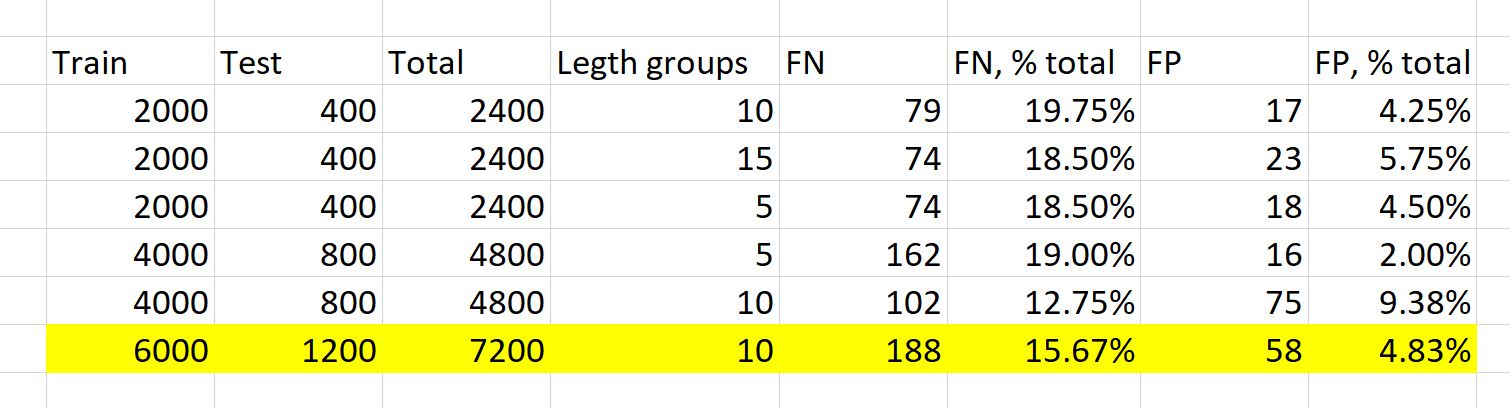}
\caption{EN-JA news items predictions with fine-tuning completed on different training dataset sizes, \textbf{\textit{gpt3.5turbo}} model}
\label{fig:en-JA-varying-training-sets}
\end{center}
\end{figure}

We ran several experiments with varying training set sizes, with results shown in Figure \ref{fig:en-JA-varying-training-sets}.

These results are interesting because although False Positive prediction does not improve with the increase of training set, in the context of the need for post-editing the False Negative category is much more interesting, because we are interested in better prediction of those segments which do NOT require post-editing. And, as we see from the experimental data, the prediction of FN improves from almost 20\% to 12\%-15\% with the increase of training set from 2000 bilingual segments to 6000 bilingual segments.

We, therefore, can recommend the training set in that range, since larger sizes of training set will be more expensive and will take significant time for models with the size of \textbf{\textit{gpt3.5turbo}}.

\subsection{Comparison of performance on different OpenAI models}
It was also interesting to see how the extra-large LLMs (xLLMs) from OpenAI, the \textbf{\textit{davinci}} and \textbf{\textit{gpt3.5turbo}} models, perform on the same task in comparison to \textbf{\textit{curie}} model we used earlier.
These three LLMs have parameter sizes around 13B, 175B, and 175B respectively.

So we used the same English-Italian data from our original experiment to compare performance on different models of the same EN-IT dataset.

\begin{figure*}[!h]
\begin{center}
\centering
\includegraphics*[width=0.99\textwidth]{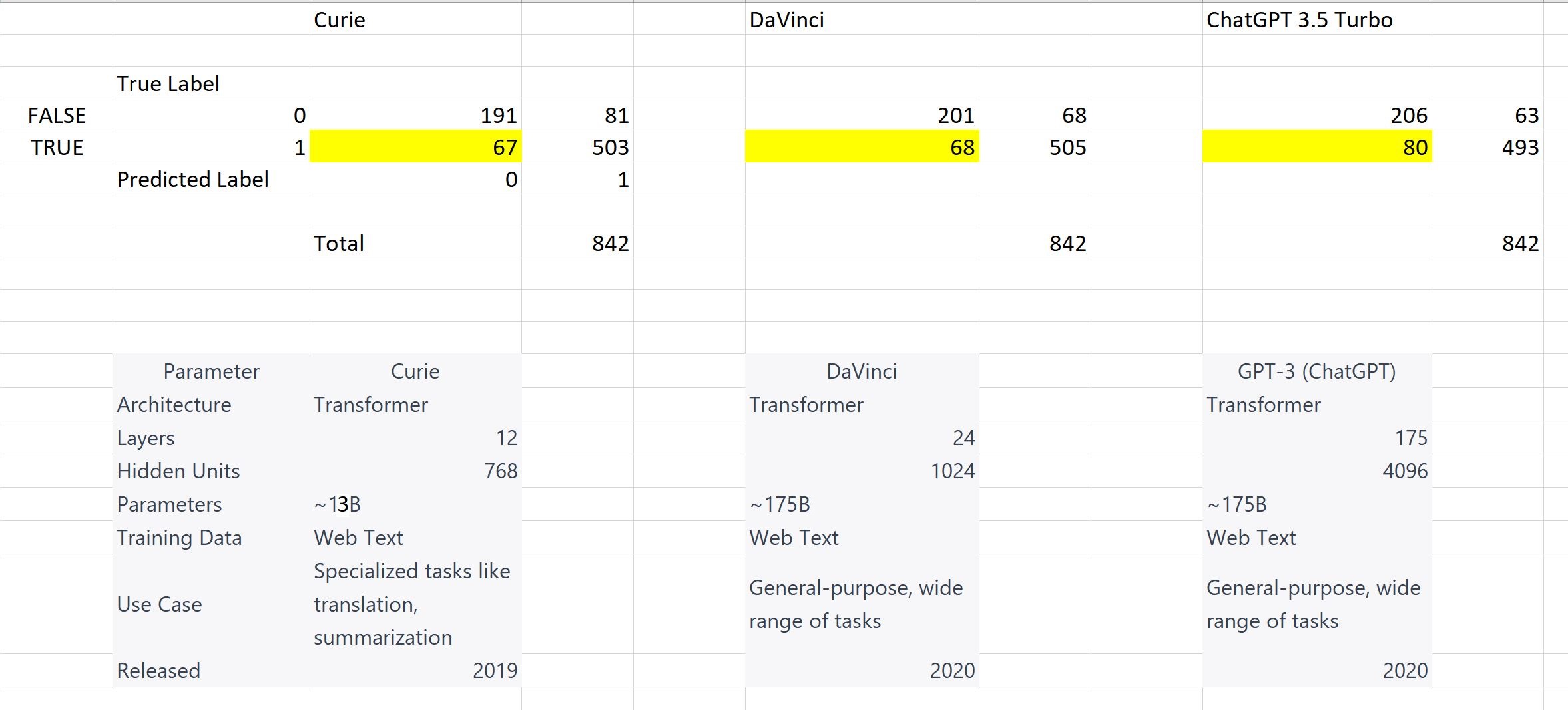}
\caption{Comparisons of three LLMs on Confusion Matrix and Parameters.}
\label{fig:curie-Davinci-chatgpt3.5-compare}
\end{center}
\end{figure*}

Figure \ref{fig:curie-Davinci-chatgpt3.5-compare} shows the comparison of these three LLMs regarding their confusion matrix and parameter sets. 
Surprisingly, their performances on predicting MT errors are very close, i.e. the larger-sized \textbf{\textit{davinci}} model and extra-large sized \textbf{\textit{gpt3.5turbo}} did not demonstrate much improvement on model classification accuracy. Their correct labels (TP+TN) are  (694, 699, 706) respectively out of 842 all labels, which results in the accuracy ratios 82.42\%, 83.02\%, and 83.85\%. 
In comparison to the much smaller \textbf{\textit{curie}} model with 12 layers of Transformer and 768 hidden units, the xLLM \textbf{\textit{gpt3.5turbo}} only achieved 1.43 points (83.85\%-82.42\%) increase of accuracy score despite using 175 layers of Transformer and 4096 hidden units.

The explanation for this may probably be found because the fine-tuning loss on this classification task drops down very quickly.

\begin{figure*}[!h]
\begin{center}
\centering
\includegraphics*[width=0.99\textwidth]{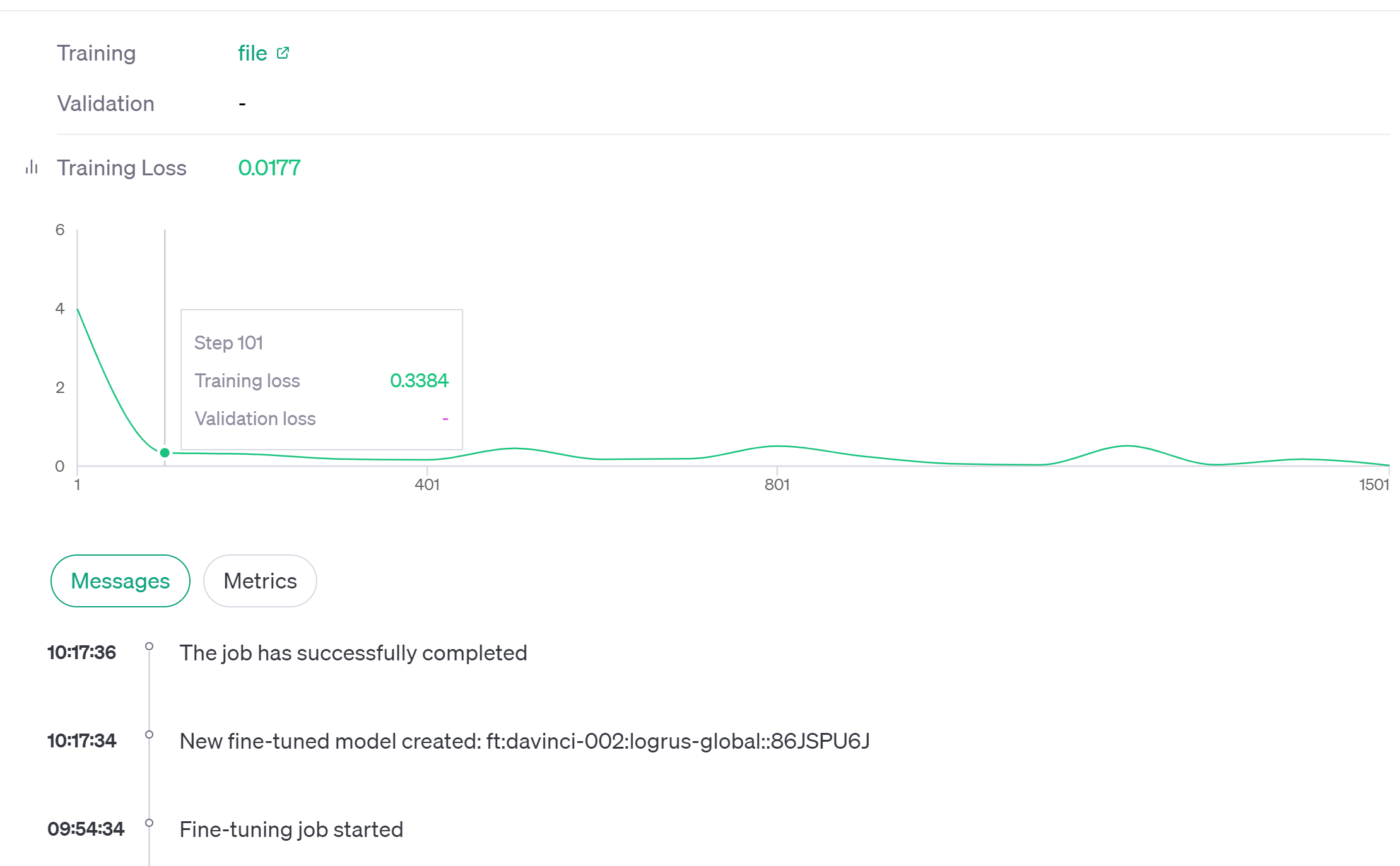}
\caption{Fine-tuning progress on \textbf{\textit{gpt3.5turbo}} model fine-tuning.}
\label{fig:chatgpt3.5-ft-loss}
\end{center}
\end{figure*}

Figure \ref{fig:chatgpt3.5-ft-loss} shows the fine-tuning loss on the \textbf{\textit{gpt3.5turbo}} model. As can be seen from this graph, only 100 steps are sufficient to bring the loss to almost zero, and then all other steps contribute very little to the classification quality improvement.

As we can see, there is no need to use larger models since results hardly improve as compared with \textbf{\textit{curie}} model.

\section{Conclusions and Future Work}
\label{sec_conclude}
In this work, to investigate the LLM's capability of predicting MT output errors, we fine-tuned GPT models via OpenAI API. 
We formulated the task as a classification challenge using prepared historical post-editing data on English-Italian and English-German for pilot studies. 
The experimental output using fine-tuned LLMB2PEN demonstrated promising results. 
We also analysed the possible solutions for addressing the error rates, i.e. whether prediction errors can be ignored and published without the review, or letting them be reviewed by the linguists at a lower rate, and how much saving can be achieved for the client who uses this process, in comparison to 100\% post-editing without using LLMB2PEN method.

In the extended experiments, we added six more language pairs including English-to-French, Japanese, Dutch, Portuguese, Turkish, and Chinese, in total resulting in eight, and summarised our findings by classifying the language pairs. 
We also compared GPT models from different sizes and the experimental results surprisingly show that the larger LLMs (\textbf{\textit{davinci}} and \textbf{\textit{gpt3.5turbo}}) do not improve the accuracy performance of much smaller \textbf{\textit{curie}} model with apparent margins but with much more cost.

In the future, we are going to work on response rate and training times to see whether the model can continue learning as \textit{being fed with more consecutive chunks of data} for the same languages, to implement an ongoing learning of prediction.
In addition, we plan to carry out the LLMB2PEN fine-tuning on other language pairs for which we have historical data. We intend to explore to what extent the model is capable of absorbing data for several languages, i.e. one fine-tuned multilingual model serving several language pairs.


To further extend this project, it will also be interesting to explore and check whether the LLMB2PEN method can help to identify human-introduced errors or translationese.






\section*{Limitations}
In this work, we reported MT QE experiments using eight language data translated from English. The positive results produced from the OpenAI models can be further enhanced by more language pairs, as well as broader domains of the corpus. 

The main limitation of the method is non-zero fine-tuning time. The fine-tuning takes about 20 minutes and therefore cannot be made continuous, which has to be done periodically, in batches. This hardly can be overcome, but deployment methods can be applied to quickly replace the older fine-tuned models with the newer ones.

\section*{Ethical Statement}
This work has no ethical concerns since we did not disclose any identifiable private user data. All experiments were carried out in a secure computing environment.

    \section*{Acknowledgements}
We thank Georg Kirchner, Globalization Technology Manager at Dell Technologies, for the valuable comments on the initial manuscript. 
LH and GN are grateful for the support from the grant “Assembling the Data Jigsaw: Powering Robust Research on the
Causes, Determinants and Outcomes of MSK Disease”. The project has been funded by the Nuffield
Foundation, but the views expressed are those of the authors and not necessarily the Foundation. 
Visit www.nuffieldfoundation.org. 
LH and GN are also supported by the grant “Integrating hospital outpatient letters into the healthcare data space” (EP/V047949/1; funder: UKRI/EPSRC).

\bibliography{eamt24,eamt24-latex-template/anthology}
\bibliographystyle{eamt24}

\end{document}